\documentclass[sigconf]{acmart}

\usepackage{booktabs}
\usepackage{colortbl}
\usepackage{graphicx}
\usepackage{array}

\AtBeginDocument{%
  }

\copyrightyear{2026}
\acmYear{2026}
\setcopyright{cc}
\setcctype{by}
\acmConference[SIGIR '26]{Proceedings of the 49th International ACM SIGIR Conference on Research and Development in Information Retrieval}{July 20--24, 2026}{Melbourne, VIC, Australia}
\acmBooktitle{Proceedings of the 49th International ACM SIGIR Conference on Research and Development in Information Retrieval (SIGIR '26), July 20--24, 2026, Melbourne, VIC, Australia}
\acmDOI{10.1145/3805712.3808445}
\acmISBN{979-8-4007-2599-9/2026/07}

\begin{document}

\title{What Gets Cited: Competitive GEO in AI Answer Engines}

\author{Rahul Vishwakarma}
\email{rahul.vishwakarma@sprinklr.com}
\affiliation{%
  \institution{Sprinklr}
  \city{Gurugram}
  \country{India}
}

\author{Shushant Kumar}
\email{shushant.k@sprinklr.com}
\affiliation{%
  \institution{Sprinklr}
  \city{Dubai}
  \country{UAE}
}

\author{Ratnesh Jamidar}
\email{ratnesh.jamidar@sprinklr.com}
\affiliation{%
  \institution{Sprinklr}
  \city{Gurugram}
  \country{India}
}

\renewcommand{\shortauthors}{Rahul Vishwakarma, Shushant Kumar, and Ratnesh Jamidar}

\begin{abstract}

AI answer engines generate answers from retrieved pages but cite only a few sources. This makes visibility depend not just on ranking, but on being cited. We study competitive Generative Engine Optimization (GEO): when two retrieved candidates compete, what makes one more likely to be cited first? We build a controlled two-document retrieval-augmented generation (RAG) testbed that \emph{injects} exactly two candidate sources into the model context and measures which source is referenced by the first citation marker in the output. Across six LLMs we execute 252{,}000 \emph{trials}, repeated paired comparisons under one factorial program over 18 content factors. In each trial the two sources differ in exactly one factor; we use brand anonymization and counterbalanced source order to separate content effects from position bias. Mixed-effects models show that topical relevance and list position are the biggest drivers of being cited first. Including explicit price information and a recent timestamp also helps consistently. Completeness and trust cues add smaller gains, while formatting-only edits have little impact. We release a reproducible evaluation protocol and a prioritized GEO checklist for practitioners, and we exercised it in an early internal pilot at Sprinklr, where teams reported positive qualitative feedback on workflow usability.

\end{abstract}

%



\maketitle

\section{Introduction}
\label{sec:introduction}

\begin{table*}[!t]
\caption{GEO Taxonomy of 18 Tested Factors with Research Backing}
\label{tab:taxonomy}
\vspace{-4mm}
\centering
\fontsize{7.75}{10}\selectfont
\renewcommand{\arraystretch}{1.4}
\setlength{\tabcolsep}{6pt}
\setlength{\aboverulesep}{0pt}
\setlength{\belowrulesep}{0pt}
\begin{tabular}{p{2.4cm}p{2.8cm}p{6cm}p{4.8cm}}
\toprule
\rowcolor[gray]{0.85}
\textbf{Category} & \textbf{Factor} & \textbf{Description} & \textbf{Supporting Evidence} \\
\midrule
\rowcolor[gray]{0.95}
Content Match & Topic Mismatch & Content discusses unrelated products instead of query-relevant items & BM25 \cite{robertson2009}, TF-IDF, semantic relevance \\
\midrule
Content Match & Keyword Gap & Content lacks key query terms present in competing sources & Hybrid search in RAG, DPR \cite{karpukhin2020} \\
\midrule
\rowcolor[gray]{0.95}
Completeness & Price Not Mentioned & Product price information absent & Decision-making research, purchasing factors \\
\midrule
Completeness & Missing Specifications & Technical specifications omitted & Information completeness standards \\
\midrule
\rowcolor[gray]{0.95}
Completeness & No Comparisons & No comparison with alternative products & AI synthesis capability unclear \\
\midrule
Trustworthiness & Hedged Language & Heavy use of uncertain qualifiers (might, possibly, could) & Linguistic hedging and certainty \cite{hyland1994} \\
\midrule
\rowcolor[gray]{0.95}
Trustworthiness & Claims With Evidence & Claims lack supporting evidence (tests, certifications) & Information credibility research \cite{metzger2013} \\
\midrule
Trustworthiness & Internal Contradictions & Content contains conflicting statements & Information consistency principles \\
\midrule
\rowcolor[gray]{0.95}
Trustworthiness & Overly Promotional & Excessively enthusiastic or sales-focused tone & AI safety training observations \\
\midrule
Readability & Content Structure & Dense paragraph format versus organized sections & Document understanding research \\
\midrule
\rowcolor[gray]{0.95}
Readability & Scattered Information & Related information dispersed throughout text & Information grouping principles \\
\midrule
Competitive Standing & Weaker Value Proposition & Less compelling benefits versus competitor & Marketing communication research \\
\midrule
\rowcolor[gray]{0.95}
Competitive Standing & Less Comprehensive & Shallower analysis versus detailed alternative & Information completeness preference \\
\midrule
Competitive Standing & Weaker Social Proof & Fewer or lower ratings/reviews & Social proof in online reviews \cite{hu2006} \\
\midrule
\rowcolor[gray]{0.95}
Competitive Standing & Lower List Position & Source appears in Position 2 versus Position 1 & \textit{Lost in the Middle} \cite{liu2024}, position bias \cite{schilcher2025} \\
\midrule
Freshness & Recent vs Old Timestamp & Content dated 2026 versus 2019 & Temporal IR research \cite{efron2011}, recency bias \\
\midrule
\rowcolor[gray]{0.95}
Freshness & No vs Old Timestamp & No timestamp versus 2019 date & Staleness signals \\
\midrule
Freshness & Recent vs No Timestamp & Recent (2026) timestamp versus no timestamp & Competing theories on timestamp absence \\
\bottomrule
\end{tabular}
\end{table*}

AI-powered search systems increasingly use retrieval-augmented generation (RAG)~\cite{lewis2020}: they retrieve candidate documents and use an LLM to synthesize an answer with selective source citations. Because only a small subset of retrieved sources are cited, citation selection becomes a visibility bottleneck. Traditional SEO focused on ranking positions~\cite{brin1998}, but in search with LLMs, content owners must also optimize for being \emph{cited} in the generated response. We follow Aggarwal et al.~\cite{aggarwal2024} in calling this \emph{Generative Engine Optimization (GEO)}.

Although GEO is gaining attention, most evaluations do not match the competitive citation setting practitioners face. Answer engines cite only a handful of sources, so a page must be not only "good enough" in isolation, but \emph{preferred} over other plausible candidates competing for the same citation slot. Aggarwal et al.~\cite{aggarwal2024} take an important first step by introducing GEO and proposing benchmark-style visibility metrics, showing that targeted rewrites can increase how much a model reflects a given source. However, these evaluations primarily quantify \emph{single-source visibility} within a fixed retrieved set and do not directly estimate \emph{citation preference}: which of two similar candidates wins a citation when they compete directly. Beyond evaluation gaps, observational analyses on live systems further complicate causal inference by mixing content effects with retrieval rank, presentation order, and other interface artifacts. Without controlled experiments that isolate individual content factors, practitioners risk misidentifying citation drivers, leading to wasted optimization effort while missing gatekeeping factors that determine whether a source is cited at all.

We study \emph{competitive} GEO using repeated head-to-head matched-pair comparisons in a simulated RAG setting where two candidates are injected as the only retrieved results. To remove two dominant confounds, we (i) counterbalance source order to separate content effects from position bias in prompts~\cite{liu2024,schilcher2025} and (ii) anonymize brands and publishers to reduce familiarity effects from pretraining. From 100 anonymized product review articles across 50 categories, we construct 1{,}440 scenarios in which two variants differ in exactly one of 18 factors spanning content match, completeness, trustworthiness, readability, competitive standing, freshness, and position. We then execute 252{,}000 \emph{trials} across six LLMs and model citation preference using logistic mixed-effects models.

We make three contributions:
\vspace{-1mm}
\begin{enumerate}
\item We propose a simple, controlled way to test GEO under competition: show an LLM two sources at a time, change only one content factor between them, anonymize brands/publishers, and swap their order to control for position bias.
\item We use this protocol at scale to measure how 18 content factors change which source is cited first across six LLMs, using mixed-effects models.
\item We translate the results into a prioritized, evidence-based checklist that teams can use to improve citation visibility in AI search.
\end{enumerate}

\vspace{-3mm}
\section{Methodology}
\label{sec:methodology}

Identifying which content attributes drive LLM citations requires controlled experiments because real-world content varies in multiple attributes simultaneously. We employ head-to-head A/B tests in an injected two-source RAG setup: for each query, the LLM receives two candidate sources (Variant A and Variant B) differing in exactly one characteristic, and generates an answer with citations. We record \emph{citation preference}: which source is cited first.

We first construct a controlled dataset that isolates individual content factors while removing brand familiarity confounds, then execute experiments at scale across six LLMs and analyze results with mixed-effects models.

\vspace{-2mm}
\subsection{Dataset Creation}
\label{sec:dataset}

\vspace{-2mm}
\begin{figure}[h]
\centering
\includegraphics[width=0.9\columnwidth]{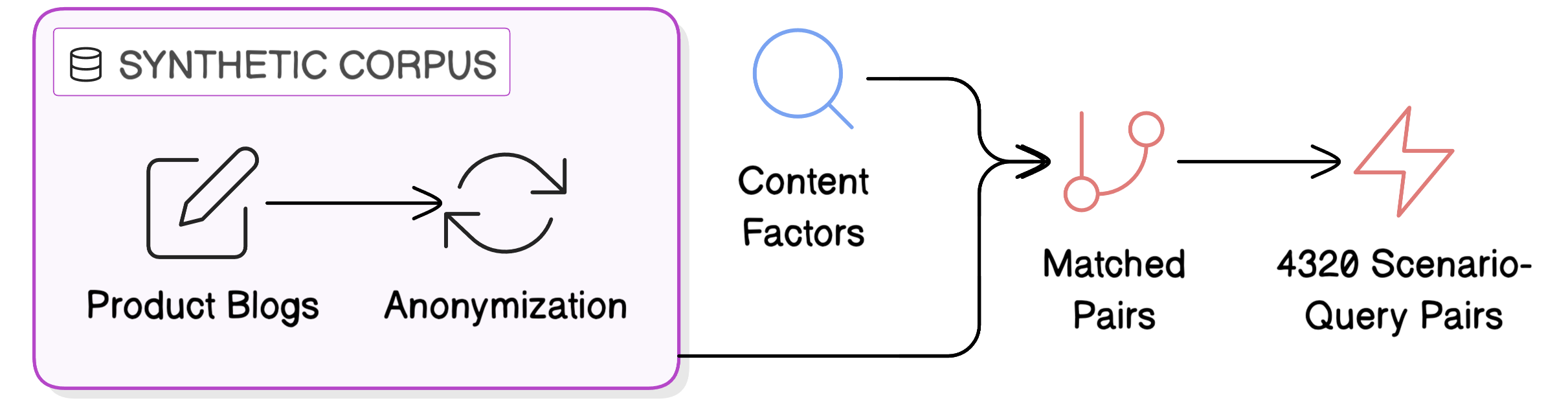}
\vspace{-3mm}
\caption{Dataset creation pipeline}
\Description{Pipeline diagram flowing left to right with five elements. A grouped box labeled Synthetic Corpus contains two elements connected by an arrow: 100 Product Blogs flows to Bias Control. Outside the group, 18 Content Factors is a separate element. Both the Synthetic Corpus group and 18 Content Factors converge via arrows into Matched Pairs, which flows to 4,320 Scenario-Query Pairs.}
\label{fig:dataset-pipeline}
\end{figure}
\vspace{-2mm}

The pipeline has four stages (Figure \ref{fig:dataset-pipeline}): define which factors to test, curate real-world seed content, anonymize brands to prevent familiarity from affecting results, and generate matched pairs that differ in exactly one factor.

\textbf{18-Factor Testing Framework.} To focus on attributes that content creators can directly control, we synthesized a taxonomy of 18 factors from information quality frameworks \cite{wang1996}, RAG systems research \cite{lewis2020}, position bias studies \cite{liu2024,schilcher2025}, and information retrieval (IR) principles \cite{karpukhin2020,reichman2024,robertson2009}. The factors span Content Match, Completeness, Trustworthiness, Readability, Competitive Standing, and Freshness, plus a position-bias manipulation (Lower List Position). Table \ref{tab:taxonomy} presents the complete list.

\textbf{Seed Corpus Curation.}
We selected 50 diverse B2C product categories (consumer technology, home goods, fitness equipment, etc.) and used GPT-4o with web-search support to curate two representative product review blog posts per category, 100 in total, capturing 2026 market prices, key technical specifications, and publication timestamps.

\textbf{Familiarity Bias Control.} 
LLMs can favor well-known entities based on pretraining exposure \cite{kamruzzaman2024brand}. To remove this confound, we used GPT-4o to replace brand names, product models, and publisher names with fictional aliases across all 100 blogs, preserving factual content (prices, specifications, and timestamps). This ensures citation outcomes reflect content characteristics, not brand recognition.

\textbf{Matched-Pair Design.} To isolate each factor's causal effect, for each of the 18 factor-wise hypotheses we randomly selected 20 blogs and used GPT-4o to generate 4 scenarios per blog, producing 80 scenarios per factor and 1,440 across all factors. Each scenario pairs two variants that differ in exactly one factor while matching in all facts, prices, specifications, and length ($\pm$5\%). To ensure results generalize across query wording, we generated 3 query paraphrases for each of the 1,440 base scenarios, yielding 4,320 scenario-query instances across all 18 tests. All three authors independently reviewed a stratified sample of 300 scenarios (21\% of 1,440), checking anonymization completeness, single-factor isolation, factual consistency between variants, and length parity. No brand leakage or unintended factor differences were found.

\smallskip
\noindent\fbox{\begin{minipage}{0.975\columnwidth}
\textbf{Variant A (Confident):} \textit{The CleanBot Aroma Pro X3 delivers exceptional cleaning with 30,000 Pa suction and 99.99\% germ elimination.}

\textbf{Variant B (Hedged):} \textit{The CleanBot Aroma Pro X3 might possibly deliver cleaning with what could be around 30,000 Pa suction...}
\end{minipage}}

\subsection{Execution}

We first describe how trials were run with counterbalanced ordering across six LLMs, then how we modeled citation outcomes.

\textbf{Why Two Sources?} A pairwise design yields a clean forced choice that isolates one content factor at a time. It also makes position control tractable: with two sources there are only two orders, whereas \emph{fully counterbalancing} $k$ sources requires $k!$ permutations (6 for $k{=}3$, 24 for $k{=}4$), quickly multiplying executions (API/compute budget) and complicating attribution.

\textbf{Experimental Execution.} Each trial follows the usual turn order: \textbf{system} message, \textbf{user} message, a \texttt{web\_search} \textbf{tool call}, the \textbf{tool response}, then the model answers with citations. We use one fixed system prompt for every trial:

\smallskip
\noindent\fbox{\begin{minipage}{0.975\columnwidth}
\small\textit{``You are a helpful assistant. You MUST use the web\_search tool to find information before answering any query. Always cite the exact URLs from the search results in your response. Do not use any other sources or prior knowledge.''}
\end{minipage}}
\smallskip

\noindent The user message carries only the scenario question. Next we insert the \texttt{web\_search} \textbf{tool call} and \textbf{tool response}. In a live product the model usually writes the tool call and a backend runs the search to build the response. Here we supply both: the call carries a query string chosen when we built the plan for that scenario-query pair, and we reuse that same call (including the query) on every repeated run of the pair. The response always lists exactly two variant sources, each with a title, a URL, and the full text, in counterbalanced order. We never call a search engine. The model then answers with citations and sees only this pair. All six APIs follow the same pattern. Vendors differ only in how they format tool calls, not in the text the model reads.

\vspace{-2mm}
\begin{figure}[h]
\centering
\includegraphics[width=0.9\columnwidth]{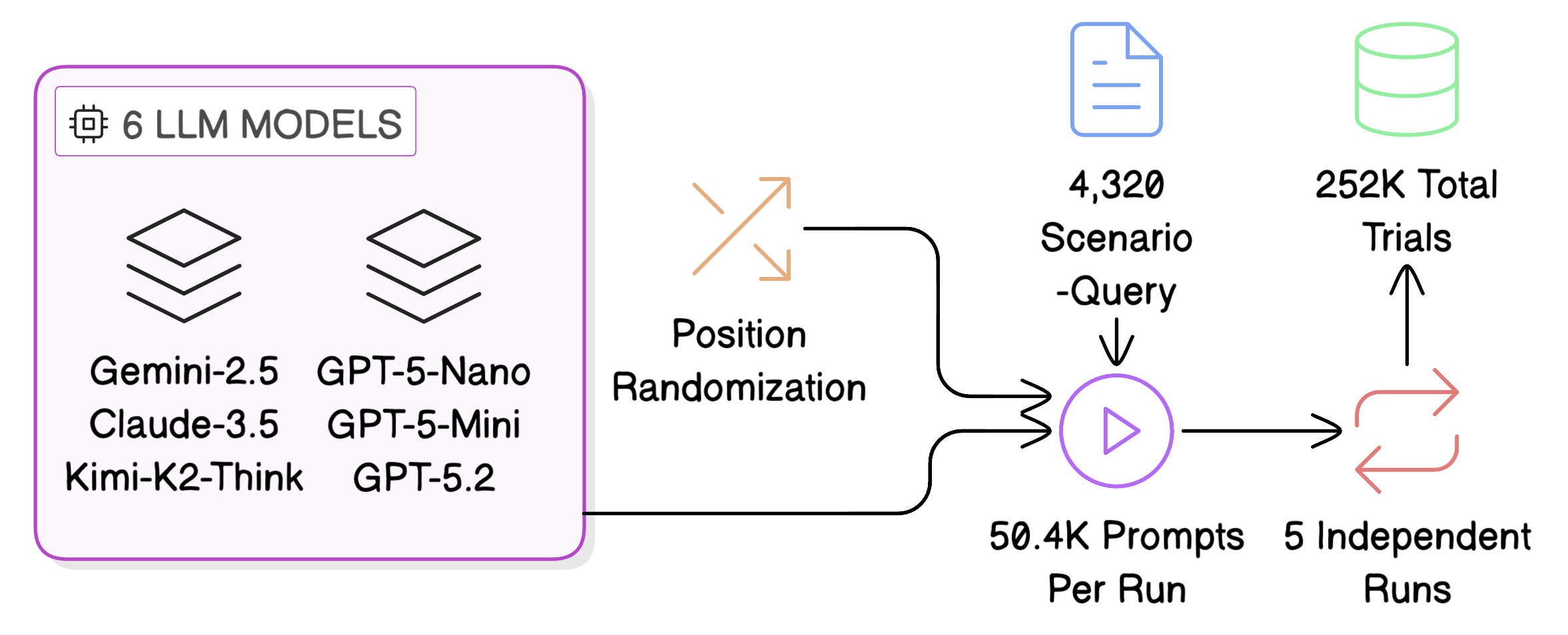}
\vspace{-3mm}
\caption{Experimental execution scale}
\Description{Flow diagram with six elements connected by arrows. A grouped box labeled 6 LLM Models lists Gemini-2.5, GPT-5-Nano, Claude-3.5, GPT-5-Mini, Kimi-K2-Thinking, and GPT-5.2. Three inputs converge into a central execution node: the 6 LLM Models group, Position Randomization, and 4,320 Scenario-Queries. The execution node is labeled 50.4K Prompts Per Run, which flows to 5 Independent Runs, then to 252K Total Trials.}
\label{fig:execution-pipeline}
\end{figure}
\vspace{-2mm}

Since LLMs can favor earlier-listed sources \cite{liu2024,schilcher2025}, we counterbalanced source orders ([A,B] and [B,A]) for 17 content-quality factors, but not for Lower List Position, where order is the treatment. We ran each combination of scenario, query, and order 5 times for statistical stability. This yielded 2{,}400 trials per factor for 17 factors, plus 1{,}200 for the position factor, totaling 42{,}000 trials per model. Six LLMs (Gemini-2.5-Flash, GPT-5-Nano, GPT-5-Mini, GPT-5.2, Claude-3.5-Sonnet, Kimi-K2-Thinking) produced 252{,}000 trials in total (Figure \ref{fig:execution-pipeline}). We held each model's output-length limit and, where the API allows it, sampling temperature fixed; across trials only the scenario question, the two source texts, their order in the tool response, and stochastic variation across repeated runs changed.

\noindent Table~\ref{tab:effect-sizes} reports fitted odds ratios by factor and model for the trials just described (same fixed system prompt as in the framed quote).

\begin{table}[!t]
\caption{Quantitative Effect Sizes: Odds Ratios Across Models}
\label{tab:effect-sizes}
\vspace{-3mm}
\centering
\fontsize{7.75}{10}\selectfont
\renewcommand{\arraystretch}{1.3}
\setlength{\tabcolsep}{2pt}
\setlength{\aboverulesep}{0pt}
\setlength{\belowrulesep}{0pt}
\begin{tabular}{p{3.05cm}|p{0.85cm}p{0.85cm}p{0.75cm}|p{0.7cm}p{0.7cm}p{0.6cm}}
\toprule
\rowcolor[gray]{0.85}
\textbf{Factor (A vs B)} & \multicolumn{3}{c}{\textbf{Diverse Providers}} & \multicolumn{3}{c}{\textbf{GPT Models}} \\
\rowcolor[gray]{0.95}
  & \textbf{Gemini 2.5 Flash} & \textbf{Claude 3.5 Sonnet} & \textbf{Kimi K2 Thinking} & \textbf{GPT 5 Nano} & \textbf{GPT 5 Mini} & \textbf{GPT 5.2} \\
\midrule
\rowcolor[gray]{0.95}
On-Topic vs Off-Topic & \textbf{$>$10k} & \textbf{$>$10k} & \textbf{$>$10k} & \textbf{221}$^\dagger$ & \textbf{$>$10k} & \textbf{$>$10k} \\
\midrule
Query Terms vs Missing & 9.41 & 17.0$^\dagger$ & \textbf{16.4} & \textbf{5.99} & \textbf{14.4} & \textbf{40.0} \\
\midrule
\rowcolor[gray]{0.95}
Price vs No Price & \textbf{$>$10k} & \textbf{$>$10k} & \textbf{36.1} & \textbf{7.82} & \textbf{6.26} & \textbf{30.4} \\
\midrule
Specs vs No Specs & 8.63 & \textbf{$>$10k} & \textbf{238} & \textbf{11.5} & \textbf{15.4} & \textbf{243} \\
\midrule
\rowcolor[gray]{0.95}
With vs No Comparisons & \textbf{7.45}* & \textbf{4.55}* & \textbf{4.72} & \textbf{2.14} & 1.78* & 1.61 \\
\midrule
Confident vs Hedged & \textbf{599} & 754 & \textbf{10.6} & \textbf{2.67}* & \textbf{5.44} & \textbf{4.75} \\
\midrule
\rowcolor[gray]{0.95}
Evidence vs No Evidence & 8.00* & \textbf{$>$10k} & \textbf{46.3} & \textbf{2.73} & \textbf{5.91} & 2.09 \\
\midrule
Consistent vs Contradictory & 2.81 & \textbf{2.81} & \textbf{2.72} & \textbf{2.19} & \textbf{4.09} & 1.74 \\
\midrule
\rowcolor[gray]{0.95}
Neutral vs Promotional & 1.45* & 362* & 2.02 & 1.31 & \textbf{1.81} & \textbf{2.08}* \\
\midrule
Structured vs Dense & 1.68 & 1.03$^\dagger$ & 0.79 & 0.90 & 0.78 & 1.25 \\
\midrule
\rowcolor[gray]{0.95}
Organized vs Scattered & 2.21 & 3.87 & 2.49 & 1.19 & 1.13 & 1.57 \\
\midrule
Strong vs Weak Value Prop & 5.22 & 5.66 & \textbf{7.24} & \textbf{2.79} & \textbf{2.68} & 1.53* \\
\midrule
\rowcolor[gray]{0.95}
Deep vs Shallow Coverage & 1,480$^\dagger$ & \textbf{$>$10k} & \textbf{132} & \textbf{5.33} & \textbf{19.5} & \textbf{3.98} \\
\midrule
Strong vs Weak Social Proof & 6.61* & $>$10k & \textbf{25.4} & \textbf{3.13} & \textbf{4.78} & 2.14 \\
\midrule
\rowcolor[gray]{0.95}
Position 1 vs 2 & \textbf{$>$10k}* & \textbf{$>$10k} & \textbf{$>$10k} & \textbf{2,002} & \textbf{1,795} & \textbf{$>$10k} \\
\midrule
Recent vs Old Timestamp & \textbf{$>$10k} & \textbf{$>$10k} & \textbf{68.7} & \textbf{14.4} & \textbf{1,494} & \textbf{$>$10k} \\
\midrule
\rowcolor[gray]{0.95}
No vs Old Timestamp & 2.32 & 2.28 & \textbf{1.33}* & 1.31 & 1.55 & \textbf{1.48}* \\
\midrule
Recent vs No Timestamp & 1.99 & 13.0 & \textbf{3.67} & 1.15 & \textbf{3.53} & \textbf{1.99} \\
\bottomrule
\end{tabular}

\smallskip
\noindent\textbf{\textit{Bold}} = statistically significant ($p < 0.05$). OR $> 1$ favors variant A.

\noindent\textbf{\textit{Convergence warnings:}} *Degenerate Hessian, $^\dagger$Singular fit.

\noindent\textbf{\textit{Effect Size Categories:}} Very strong (OR $>$ 100), Strong (OR 10--100), Moderate (OR 3--10), Weak (OR 1.5--3), Negligible (OR $<$ 1.5)
\vspace{-3mm}
\end{table}

\begin{figure*}[h]
\centering
\includegraphics[width=0.95\textwidth]{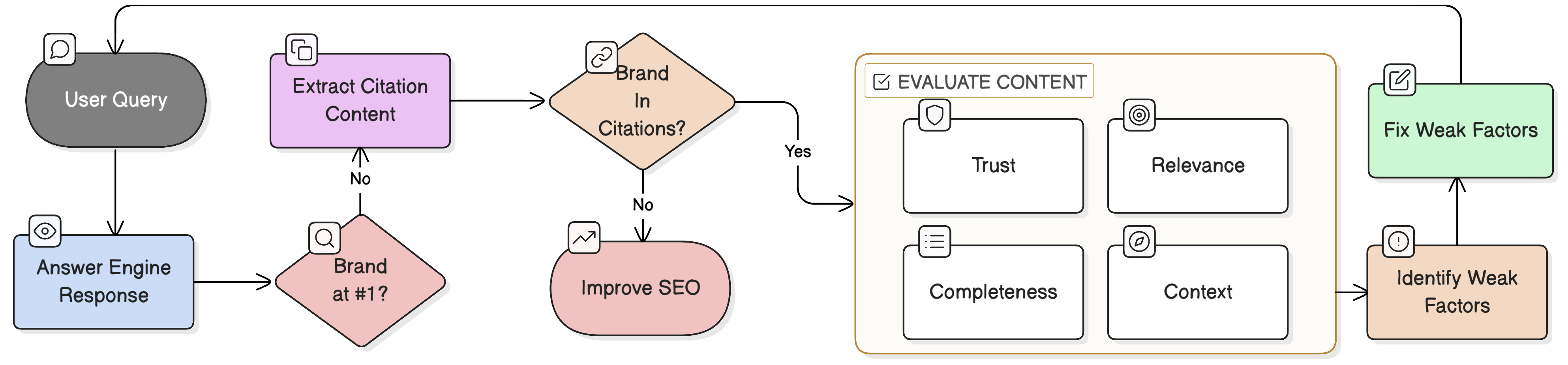}
\vspace{-4mm}
\caption{Content optimization workflow}
\Description{Flowchart with nine elements including two decision diamonds, connected by directional arrows. Start: User Query leads to Answer Engine Response, then decision diamond 'Brand at number 1?'. If No, Extract Citation Content leads to decision diamond 'Brand In Citations?'. If Yes, Evaluate Content across four categories (Trust, Completeness, Relevance, Context), then Identify Weak Factors, then Fix Weak Factors, which loops back to User Query. If brand not in citations, flow leads to Improve SEO.}
\label{fig:geo-workflow}
\end{figure*}

\textbf{Statistical Analysis.} From each LLM response, we extracted which source was cited first, giving a binary outcome $Y_i = 1$ if Variant~A is cited first in trial $i$, and 0 otherwise. To avoid pseudoreplication \cite{hurlbert1984} from repeated runs nested within position orders within scenarios, we applied logistic GLMM with nested random effects \cite{jaeger2008,bates2015}:

\vspace{-3mm}
\begin{equation}
\mathrm{logit}\big(\mathbb{P}(Y_i=1)\big) = \beta_0 + \beta_1 X_i + u_s + v_{so},
\end{equation}
\vspace{-3mm}

where $X_i$ is a centered position indicator ($+0.5$ if Variant~A is listed first, $-0.5$ otherwise) and $\beta_1$ captures presentation-order bias. We fit one model per content factor; Variant~A is the expected winner (as in Table~\ref{tab:effect-sizes}). $\beta_0$ is the fixed intercept, so $\exp(\hat{\beta}_0)$ is the odds ratio in the table after adjusting for order. $u_s$ and $v_{so}$ are random intercepts by scenario and by scenario-order combination, capturing correlation across repeated runs with the same content and order. For the Lower List Position test, we exclude the position covariate.

\vspace{-3mm}
\begin{equation}
\mathrm{logit}\big(\mathbb{P}(Y_i=1)\big) = \beta_0 + u_s + v_{so}.
\end{equation}
\vspace{-3mm}

We fitted all models in R using \texttt{lme4} with maximum likelihood estimation and the BOBYQA optimizer ($\alpha=0.05$).

\textbf{First link, multiple URLs, and exclusions.} Across successful runs for every model, answers contained exactly one distinct URL about 86.4\% of the time, two or more URLs about 10.5\%, and no URL about 3.1\%. The GLMMs keep only trials where the first URL in the answer exactly matches one of the two injected variants. About 3.4\% of trials have no URL or a first URL that matches neither variant; we drop those from the regression. If several URLs appear, we assign the outcome from the first URL in reading order.

\textbf{Effect Size Interpretation.} Table \ref{tab:effect-sizes} presents each of the 18 factor tests as \textit{A vs B}, where A is the expected-stronger variant, and reports Odds Ratios (OR $= \exp(\hat{\beta}_0)$) \cite{jaeger2008,szumilas2010} across all six models after controlling for position effects ($\beta_1$). Bold entries are statistically significant ($p < 0.05$), confirming that variant A reliably wins citation preference. For moderate effects, the OR summarizes strength (e.g., OR $= 5$ $\approx$ fivefold higher odds of being cited first). When preference is nearly deterministic, estimated ORs become very large under quasi-separation \cite{fijorek2012}; we report them as $>$10k where applicable and treat them as indicating a \emph{decisive} win for variant~A rather than a finely resolved numeric ratio. Non-bold ORs lack reliable evidence of an effect.

\vspace{-3mm}
\section{Results and Discussion}
\label{sec:results}

We organize findings by strength of cross-model agreement: factors that consistently drive citations, factors without reliable effects, and how models differ in sensitivity.

\textbf{Gatekeepers and Differentiators.} Eleven of 18 factors (61\%) reached significance in 4+ models, forming a clear hierarchy. At the top, four gatekeepers were unanimous across all six models with large effects (OR $>$ 100): Topic Mismatch, Price Not Mentioned, Recent vs Old Timestamp, and Lower List Position. Failing on any one can eliminate citation odds regardless of other content strengths. Once these prerequisites are met, seven additional factors provide secondary differentiation (OR 2.1--243): completeness (Missing Specifications, Less Comprehensive), trust (Hedged Language, Claims With Evidence, Internal Contradictions), and competitive positioning (Keyword Gap, No Comparisons).

\textbf{Factors Without Consistent Effects.} Seven factors (39\%) had weak or no effects. Formatting choices (Content Structure, Scattered Information) had no impact, suggesting LLMs parse content regardless of visual organization. The remaining five (Overly Promotional, Weaker Value Proposition, Weaker Social Proof, No vs Old Timestamp, Recent vs No Timestamp) reached significance in only two or three of six models, insufficient for consensus.

\textbf{Model Behavior Patterns.} Models varied substantially in content sensitivity. Kimi-K2 showed the highest sensitivity (83\% of factors significant), followed by the GPT family, while Claude-3.5 (50\%) and Gemini-2.5 (33\%) were more selective. Models also differed in response magnitude: Gemini and Claude exhibited categorical patterns, with 67--78\% of their significant factors producing extremely large effects (OR $>$ 10,000), suggesting binary decision boundaries. Despite these differences, all six models agreed on the four gatekeeper factors, pointing to universal citation drivers independent of model provider or training. Within the GPT family, all three models showed consistent behavior across different scales, suggesting citation patterns are shaped by architecture rather than model size.

\vspace{-3mm}
\section{Practical Implications}
\label{sec:practical-implications}

Figure \ref{fig:geo-workflow} operationalizes our findings into a practitioner-facing workflow. Given an AI answer engine response, the workflow first checks whether the brand is the top recommendation. If not, it extracts the sources the AI cited and checks whether the brand's content appears among them. If the brand's content is cited but the AI still did not recommend it first, the issue is content quality, so the workflow evaluates it against our 11-factor taxonomy (factors significant in at least four of six models) organized into four categories: Trust, Completeness, Relevance, and Context. If the brand is absent from citations entirely, the bottleneck is retrieval and the action is to improve SEO. The content evaluation produces a ranked list of weak factors with effect sizes, guiding what to fix first.

\textbf{Implementation Strategy.} The workflow applies both reactively, when citation visibility drops, and proactively for product launches or periodic content refreshes. For owned content, surface core topic terms early, add explicit price and key specs, include comparisons, keep dates current, and replace hedging with evidence-backed claims. For earned or third-party content, provide missing data to reviewers, enable independent testing, or support side-by-side comparison pieces. SEO teams drive retrieval rank, content/product teams own on-page fixes, and PR/partnerships own third-party outreach.

\textbf{Prioritization and Quick Wins.} Start with gatekeepers (topic match, price, recency, position) because failing on any one can eliminate citation odds. Quick wins are usually editorial: add concrete pricing, update timestamps, and close keyword gaps. Once the baseline is met, invest in differentiators such as spec tables, comprehensive comparisons, and competitive positioning. Formatting changes showed minimal return and can be deprioritized.

\smallskip
\noindent\fbox{\begin{minipage}{0.975\columnwidth}
\textbf{Current Content:} The Zephyr Pulse 2 fitness tracker features 8-9 day battery life and offline maps with turn-by-turn navigation. Contact us for pricing details.

\textbf{Issues Detected:} Price Not Mentioned, Missing Specifications, Weaker Value Proposition
\vspace{-1mm}
\end{minipage}}
\vspace{-2mm}

\section{Limitations and Future Work}
\label{sec:limitations}

\textbf{Beyond two injected documents.} Production RAG often retrieves five to ten or more pages, so real citation pools are larger than our testbed. We nevertheless injected \emph{exactly two} candidates because the study targets \emph{factor-level} attribution: we ask which single content attributes move first-citation preference when everything else is held fixed. Pairwise variants that differ in only one factor (with counterbalanced list order) support that identification. Each additional retrieved page would introduce simultaneous changes in overlap, redundancy, and many quality cues, so citation shifts could no longer be traced to one lever. The estimates are therefore pairwise preferences over a controlled slate, not full multi-document competition. Extending the same factorial discipline to larger candidate sets explodes conditions and orderings; we leave that extension to future work.

\textbf{Anonymization and brand or domain trust.} We anonymized brands and publishers so citation choices reflect the text, not a famous name or a well-known site the model may have seen often. Production systems may still favor trusted domains or strong brands when they pick sources. A natural complement to this work is a second factorial layer that holds our pairwise structure but swaps anonymized entities for matched real brands or domains, quantifying how much residual preference remains after the content signals we already isolate.

\textbf{LLM involvement in the corpus.} GPT-4o was used to generate seeds, perform anonymization, and create paired rewrites. The process of building the corpus is distinct from how citation preferences are evaluated. In each run, two injected segments are presented under a single protocol, and we record which of the two is preferred. This setup should not be interpreted as a judgment of the model's overall writing quality. To further ensure reliability, we manually reviewed a subset of the data. As detailed in Section~\ref{sec:dataset}, three authors conducted a stratified review of 300 scenarios, focusing on isolation of variables, alignment, and length parity, thereby reducing the likelihood that unintended differences affect the intended comparison.

\vspace{-3mm}
\section{Conclusion}
\label{sec:conclusion}

Our 252{,}000-trial study shows that LLM citation preference follows a clear hierarchy: four gatekeeper factors (topic match, price, recency, position) dominate across all six models, completeness and trust cues add secondary gains, and formatting has negligible impact. Where traditional SEO targets ranking, these findings guide what content signals win citations in AI search. We packaged them into a diagnostic workflow and validated it through a beta deployment at Sprinklr. Future work will extend to larger candidate sets, multiple citation slots, and end-to-end retrieval and recommendation pipelines.

\section*{Presenter Bio}

\textbf{Shushant Kumar} is an Associate Director of Engineering at Sprinklr, working on generative AI and agentic systems at scale. He holds an M.S. in Computer Science from Georgia Institute of Technology. His interests include retrieval-augmented generation, multimodal AI, and productionizing large-scale machine learning systems.

\bibliographystyle{ACM-Reference-Format}
\bibliography{GEO_ACM_SIGIR}

\end{document}